\documentclass[10pt,twocolumn,letterpaper]{article}

\usepackage{iccv}
\usepackage{times}
\usepackage{epsfig}
\usepackage{graphicx}
\usepackage{amsmath}
\usepackage{amssymb}
\usepackage{algorithm}
\usepackage{algorithmicx}
\usepackage{algpseudocode}
\usepackage{caption}
\usepackage{colortbl}
\usepackage{color}
\usepackage[pagebackref=true,breaklinks=true,letterpaper=true,colorlinks,bookmarks=false]{hyperref}
\usepackage{cleveref}
\usepackage{multirow}

\iccvfinalcopy % *** Uncomment this line for the final submission

\def\iccvPaperID{9589} % *** Enter the ICCV Paper ID here

\ificcvfinal\fi

\begin{document}

%-------------------------------------------------------------------------
% TITLE
%-------------------------------------------------------------------------
\title{Incremental Object Detection with CLIP}

%-------------------------------------------------------------------------
% AUTHOR
%-------------------------------------------------------------------------

% \and
\author{Yupeng He
\and
Ziyue Huang
\and
Qingjie Liu 
\and 
Yunhong Wang 
}
% Yupeng He, Ziyue Huang, Qingjie Liu, Yunhong Wang

\maketitle
\ificcvfinal\thispagestyle{empty}\fi

%-------------------------------------------------------------------------
% ABSTRACT
%-------------------------------------------------------------------------
\begin{abstract}

In the incremental detection task, unlike the incremental classification task, data ambiguity exists due to the possibility of an image having different labeled bounding boxes in multiple continuous learning stages. This phenomenon often impairs the model's ability to learn new classes. However, the forward compatibility of the model is less considered in existing work, which hinders the model's suitability for incremental learning. To overcome this obstacle, we propose to use a language-visual model such as CLIP to generate text feature embeddings for different class sets, which enhances the feature space globally. We then employ the broad classes to replace the unavailable novel classes in the early learning stage to simulate the actual incremental scenario. Finally, we use the CLIP image encoder to identify potential objects in the proposals, which are classified into the background by the model. We modify the background labels of those proposals to known classes and add the boxes to the training set to alleviate the problem of data ambiguity. We evaluate our approach on various incremental learning settings on the PASCAL VOC 2007 dataset, and our approach outperforms state-of-the-art methods, particularly for the new classes.

\end{abstract}

%-------------------------------------------------------------------------
% 1. Introducion
%-------------------------------------------------------------------------
\section{Introduction}

% 介绍增量学习概念，引入目标检测在增量学习上的必要性。
The visual system of the real world is inherently incremental, as humans need to learn new knowledge through observation and integrate it into their existing visual knowledge system. 
Despite the brilliant achievements made in object detection with deep learning, existing object detectors still struggle to adapt well to incremental learning scenarios, primarily due to issues such as catastrophic forgetting \cite{goodfellow2013empirical, kirkpatrick2017overcoming, li2017learning}. 

% 引入增量学习研究，介绍现有研究的缺陷
In recent years, researchers have made significant progress in the field of incremental learning, which allows for the easy transplantation and optimization of various approaches, leading to impressive results \cite{rebuffi2017icarl, zhu2022self, zenke2017continual, chaudhry2018riemannian, dhar2019learning, hou2019learning, dong2021bridging}. 
However, most studies focus on improving the backward compatibility of the model through techniques such as knowledge distillation \cite{hinton2015distilling} and sampler rehearsal \cite{masana2022class}. 
Few researchers have addressed the issue of forward compatibility \cite{zhou2022forward} in the incremental object detection tasks, which requires future novel categories to be easily incorporated into the current model based on the current stage data. 

% 解释什么是Backward，什么是forward，并指出检测增量特有的挑战
In incremental learning scenarios, object detection tasks may experience the phenomenon of data ambiguity \cite{shmelkov2017incremental}, leading to compatibility issues for the current model with both historical and future data, referred to as backward compatibility and forward compatibility respectively. 
Data ambiguity comes from the task requirements of the incremental detection task itself. 
In a given stage, images not only contain objects of the current stage's class but also objects from previously learned classes and potential new classes. 
Previously learned class and new class samples will be incorrectly treated as negative samples during training, resulting in compatibility issues.
As a result, the incremental detection task is more challenging than the incremental classification task, and it is hard to transfer a method from incremental classification to incremental detection. 

To address the problem of data ambiguity and forward compatibility of the detection model, we propose a method to achieves \textbf{I}ncremental \textbf{O}bject \textbf{D}etection with \textbf{C}LIP, named IODC. 
We use the zero-shot capability of vision-language pre-trained model (i.e., CLIP \cite{radford2021learning} in our work) to capture incoming new classes, and design the relevant modules to enable the growable capability of the model. 
Specifically, our approach involves enabling the model to create a language space more appropriate for new classes (Sec. \ref{sec:3.2}), learning better language space with more broad classes in the early stage and using category mapping to complete knowledge transfer (Sec. \ref{sec:3.3}), and identifying potential unknown objects in the image (Sec. \ref{sec:3.4}). 
The results demonstrate that our approach has effectively improved the model's incremental capability, particularly for learning new classes, outperforming other state-of-the-art methods.

The main contributions of our work are as follows:
\begin{itemize}
   \item We propose to add more nonexistent categories in the initial stage of incremental learning and utilize the CLIP text encoder to generate a comprehensive language space. 
   \item We use the CLIP to identify potential objects in the image to alleviate the data ambiguity problem in incremental detection tasks, which improves the model's learning ability for new classes.
   \item We utilize CLIP's multimodal feature alignment capability to identify potential objects, alleviating the data ambiguity issue in incremental learning, and improving the model's forward compatibility. 
   \item Compared with incremental detection and open world detection \cite{joseph2021towards, gupta2022ow}, our method performs better in the incremental detection task, especially improving the accuracy of new classes.
\end{itemize}

%-------------------------------------------------------------------------
% 2. Related work
%-------------------------------------------------------------------------
\section{Related work}

%-------------------------------------------------------------------------
% Incremental Learning
%-------------------------------------------------------------------------

\textbf{Incremental Learning.} In the field of image classification, there has been significant progress in developing incremental classification methods that address catastrophic forgetting, including BiC \cite{wu2019large}, PASS \cite{zhu2021prototype}, CwD \cite{shi2022mimicking}, FACT \cite{zhou2022forward}, ect. 
However, the incremental detection task has received less attention \cite{shmelkov2017incremental, perez2020incremental, yang2022multi}, particularly regarding how to improve the forward compatibility \cite{zhou2022forward} of the model to make it better suited for future tasks. Most existing work has focused on improving backward compatibility, which allows models to retain their old knowledge. For example, IODML \cite{joseph2021incremental} uses meta-learning to mitigate deviation, while ERD \cite{feng2022overcoming} leverages elastic distillation to enhance classification and regression. ELI \cite{joseph2022energy} monitors different tasks using an energy-based alignment method. While these methods are effective at consolidating previous knowledge, they do not fully consider forward compatibility, which limits the model's growth and adaptability.

%-------------------------------------------------------------------------
% Open World Object Detection
%-------------------------------------------------------------------------

\textbf{Open World Object Detection.} For traditional detection models like Faster R-CNN \cite{ren2015faster}, YOLO \cite{redmon2016you}, SSD \cite{liu2016ssd}, and RetinaNet \cite{lin2017focal}, predicting new categories is a challenge. They are unable to predict what categories may exist after initial learning. The Open World Object Detection \cite{dhamija2020overlooked} provides a solution. By using an energy-based unknown class identifier to find probable objects in the real world, the ORE \cite{joseph2021towards} enhances the model's incremental capabilities. OPENDET2 \cite{han2022expanding} detects unknown classes by expanding low-density regions and learning unknown class heads. OW-DETR \cite{gupta2022ow} explicitly encodes multi-scale contextual information and can better discriminate between unknown objects and background. However, a single undefined unknown class is created by combining all recognized unknown categories. While the early separation of unknown and background does help with incremental learning, the unknown class itself can not be divided into sub-categories, making it difficult for the model to distinguish between different object features.

%-------------------------------------------------------------------------
% Open-vocabulary Object Detection
%-------------------------------------------------------------------------

\textbf{Open-vocabulary Object Detection.} Inspired by the remarkable success of CLIP \cite{radford2021learning}, researchers have increasingly recognized the potential of language-visual models, especially their remarkable zero-shot capabilities. However, training such models typically requires large amounts of data, especially datasets containing text annotations. ViLD \cite{gu2021open} migrates CLIP to the detection task so that it can detect objects of any category. GLIP \cite{li2022grounded} has focused on training the model by emphasizing the importance of each word, thus leveraging multimodality to enhance its generalization ability. These models have highlighted the benefits of language supervision and the rich semantics inherent in class names. In this work, we aim to incorporate these benefits into traditional visual models in a flexible way, with the ultimate goal of enhancing incremental detection performance.

%-------------------------------------------------------------------------
% F1：框架图
%-------------------------------------------------------------------------
\begin{figure*}
   \begin{center}
   %\fbox{\rule{0pt}{2in} \rule{.9\linewidth}{0pt}}
   \includegraphics[width=17.5cm]{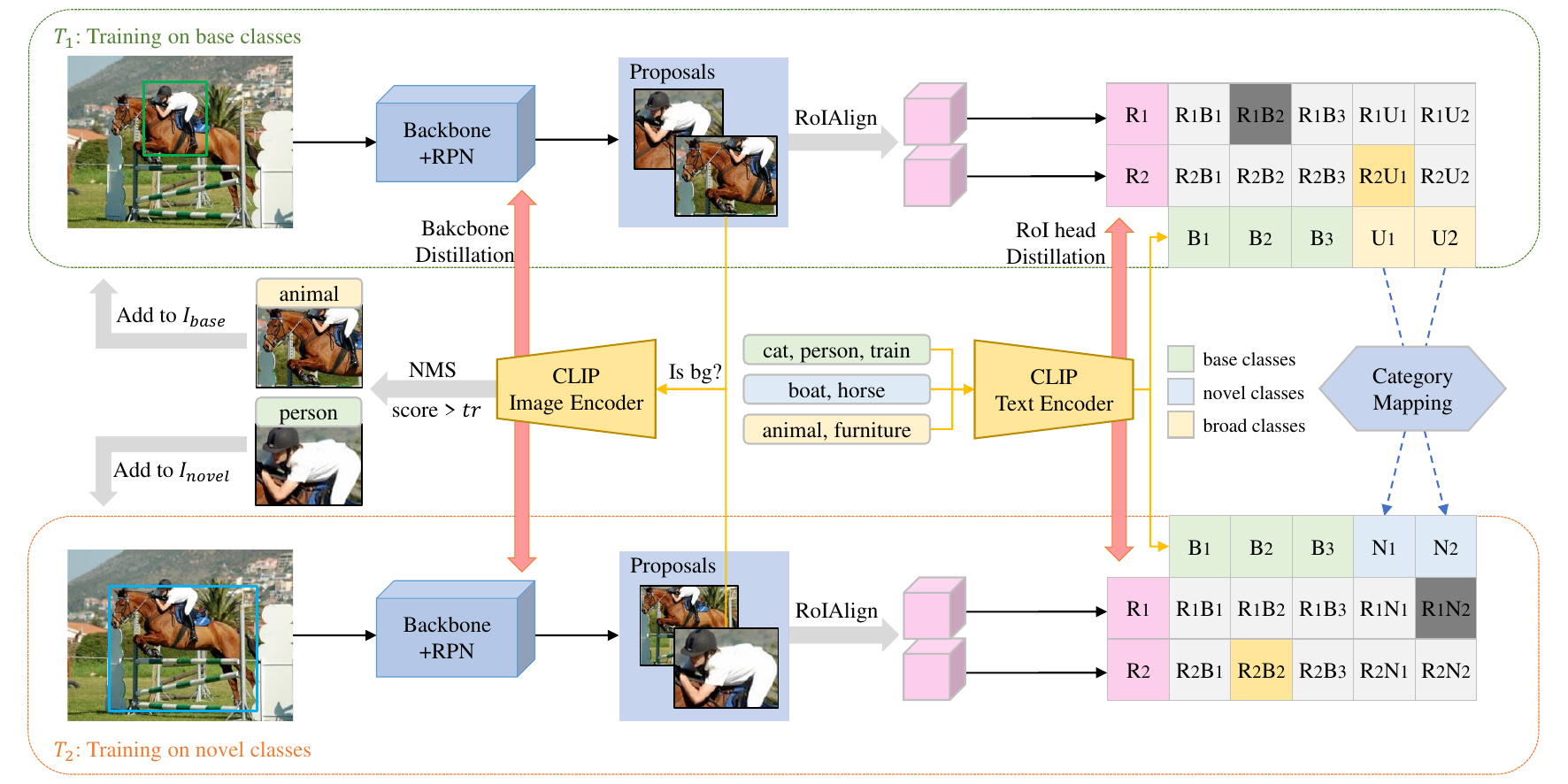} 
   \end{center}
   \caption{Approach Overview: We trained the model in $T_1$ and $T_2$ for incremental. The labeled boxes obtained at $T_1$ and $T_2$ stages are different. (a) CLIP Text Encoder: We use base and broad classes to generate text features in $T_1$, and base and novel classes in $T_2$. (b) CLIP Image Encoder: The proposals with the highest prediction of background category are sent to the CLIP image encoder for identification. We will modify the gt-label of these proposals, which are identified as the broad classes in $T_{1}$ and the base classes in $T_{2}$. R2U1 and R2B2 in yellow are used to illustrate this process. At the same time, we add these identified proposals with prediction scores higher than $tr$ to the dataset as pseudo bounding boxes after a step of NMS. When we encounter the same image in the future, we will only sample a few pseudo bounding boxes for training.}
   \label{fig:1}
\end{figure*}
% F1：框架图
%-------------------------------------------------------------------------

%-------------------------------------------------------------------------
% 3. Methodology
%-------------------------------------------------------------------------
\section{Methodology}
\label{sec:3}

To ensure robustness in different incremental settings, an incremental detection model should be designed to avoid catastrophic forgetting \cite{goodfellow2013empirical}, allowing the model to perform well on all categories. 

%-------------------------------------------------------------------------
% 3.1 Problem Formulation
%-------------------------------------------------------------------------
\subsection{Problem Formulation}
\label{sec:3.1}

As shown in Fig. \ref{fig:1}, incremental learning comprises sequential multiple tasks \cite{shmelkov2017incremental}, forming a continuous data flow. 
Each task $T_i$ introduces a group of classes denoted as $C_i$, which must not overlap with class sets introduced in other tasks, i.e., $C_i \cap  C_j= \O (i  \neq j)$. 
In each independent $T_i$ of the incremental learning process, we introduce different training data set, which can be represented as $I_i=\{(X_i, Y_i )  \vert  i=1, 2, …, N.\}$. Here, $N$ denotes the total number of images in the current traing dataset and $Y_i \in C_i$. Although the same image $X$ may appear in different $I_i$ across different learning stages $T_i$, its label $Y_i$ will always be different. 

During $T_i$ stage, the model $M_i$ is trained using only the data from the current class set $C_i$. The model $M_i$ is composed of a feature extractor $F$, a classification head $G$, and a regression head for incremental detection. 
To alleviate the problem of data ambiguity, we hope to detect potential unknown classes $C_{unk}$ in the early stages of incremental detection. 
These unknown classes are likely to appear in future novel class set $C_{novel}$, where $C_{novel} \cap C_{unk} \neq \O $ and $C_{base} \cap C_{unk} = \O $.

%-------------------------------------------------------------------------
% T1，实验大表格
%-------------------------------------------------------------------------
\begin{table*}
   \centering
   \renewcommand\arraystretch{1.3} % 调整行间距为X倍
   \renewcommand\tabcolsep{3.7pt} % 调整表格列间的宽度
% 10+10
\scriptsize % 迷你字体大小
   \begin{tabular}{l|cccccccccccccccccccc|c} 
   \hline
   \textcolor[RGB]{0, 139, 0}{\textbf{10+10 setting}} & aero & cycle & bird & boat & bottle & bus & car & cat & chair & cow & {\cellcolor[rgb]{0.882,0.882,0.882}}table & {\cellcolor[rgb]{0.882,0.882,0.882}}dog & {\cellcolor[rgb]{0.882,0.882,0.882}}horse & {\cellcolor[rgb]{0.882,0.882,0.882}}bike & {\cellcolor[rgb]{0.882,0.882,0.882}}person & {\cellcolor[rgb]{0.882,0.882,0.882}}plant & {\cellcolor[rgb]{0.882,0.882,0.882}}sheep & {\cellcolor[rgb]{0.882,0.882,0.882}}sofa & {\cellcolor[rgb]{0.882,0.882,0.882}}train & {\cellcolor[rgb]{0.882,0.882,0.882}}tv & mAP \\ 
   \hline
   Oracle 20 & 79.4 & 83.3 & 73.2 & 59.4 & 62.6 & 81.7 & 86.6 & 83.0 & 56.4 & 81.6 & {\cellcolor[rgb]{0.882,0.882,0.882}}71.9 & {\cellcolor[rgb]{0.882,0.882,0.882}}83.0 & {\cellcolor[rgb]{0.882,0.882,0.882}}85.4 & {\cellcolor[rgb]{0.882,0.882,0.882}}81.5 & {\cellcolor[rgb]{0.882,0.882,0.882}}82.7 & {\cellcolor[rgb]{0.882,0.882,0.882}}49.4 & {\cellcolor[rgb]{0.882,0.882,0.882}}74.4 & {\cellcolor[rgb]{0.882,0.882,0.882}}75.1 & {\cellcolor[rgb]{0.882,0.882,0.882}}79.6 & {\cellcolor[rgb]{0.882,0.882,0.882}}73.6 & 75.2 \\
   First 10 & 78.9 & 78.6 & 72.0 & 54.5 & 63.9 & 81.5 & 87.0 & 78.2 & 55.3 & 84.4 & {\cellcolor[rgb]{0.882,0.882,0.882}}- & {\cellcolor[rgb]{0.882,0.882,0.882}}- & {\cellcolor[rgb]{0.882,0.882,0.882}}- & {\cellcolor[rgb]{0.882,0.882,0.882}}- & {\cellcolor[rgb]{0.882,0.882,0.882}}- & {\cellcolor[rgb]{0.882,0.882,0.882}}- & {\cellcolor[rgb]{0.882,0.882,0.882}}- & {\cellcolor[rgb]{0.882,0.882,0.882}}- & {\cellcolor[rgb]{0.882,0.882,0.882}}- & {\cellcolor[rgb]{0.882,0.882,0.882}}- & 36.7 \\ 
   \hline
   ILOD \cite{shmelkov2017incremental} & 69.9 & 70.4 & 69.4 & 54.3 & 48.0 & 68.7 & 78.9 & 68.4 & 45.5 & 58.1 & {\cellcolor[rgb]{0.882,0.882,0.882}}59.7 & {\cellcolor[rgb]{0.882,0.882,0.882}}72.7 & {\cellcolor[rgb]{0.882,0.882,0.882}}73.5 & {\cellcolor[rgb]{0.882,0.882,0.882}}73.2 & {\cellcolor[rgb]{0.882,0.882,0.882}}66.3 & {\cellcolor[rgb]{0.882,0.882,0.882}}29.5 & {\cellcolor[rgb]{0.882,0.882,0.882}}63.4 & {\cellcolor[rgb]{0.882,0.882,0.882}}61.6 & {\cellcolor[rgb]{0.882,0.882,0.882}}69.3 & {\cellcolor[rgb]{0.882,0.882,0.882}}62.2 & 63.1 \\
   Faster ILOD \cite{peng2020faster} & 72.8 & 75.7 & 71.2 & 60.5 & 61.7 & 70.4 & 83.3 & 76.6 & 53.1 & 72.3 & {\cellcolor[rgb]{0.882,0.882,0.882}}36.7 & {\cellcolor[rgb]{0.882,0.882,0.882}}70.9 & {\cellcolor[rgb]{0.882,0.882,0.882}}66.8 & {\cellcolor[rgb]{0.882,0.882,0.882}}67.6 & {\cellcolor[rgb]{0.882,0.882,0.882}}66.1 & {\cellcolor[rgb]{0.882,0.882,0.882}}24.7 & {\cellcolor[rgb]{0.882,0.882,0.882}}63.1 & {\cellcolor[rgb]{0.882,0.882,0.882}}48.1 & {\cellcolor[rgb]{0.882,0.882,0.882}}57.1 & {\cellcolor[rgb]{0.882,0.882,0.882}}43.6 & 62.2 \\
   IODML \cite{joseph2021incremental} & 76.0 & 74.6 & 67.5 & 55.9 & 57.6 & 75.1 & 85.4 & 77.0 & 43.7 & 70.8 & {\cellcolor[rgb]{0.882,0.882,0.882}}60.4 & {\cellcolor[rgb]{0.882,0.882,0.882}}66.4 & {\cellcolor[rgb]{0.882,0.882,0.882}}76.0 & {\cellcolor[rgb]{0.882,0.882,0.882}}72.6 & {\cellcolor[rgb]{0.882,0.882,0.882}}74.6 & {\cellcolor[rgb]{0.882,0.882,0.882}}39.7 & {\cellcolor[rgb]{0.882,0.882,0.882}}64.0 & {\cellcolor[rgb]{0.882,0.882,0.882}}60.2 & {\cellcolor[rgb]{0.882,0.882,0.882}}68.5 & {\cellcolor[rgb]{0.882,0.882,0.882}}60.5 & 66.3 \\
   ORE \cite{joseph2021towards} & 63.5 & 70.9 & 58.9 & 42.9 & 34.1 & 76.2 & 80.7 & 76.3 & 34.1 & 66.1 & {\cellcolor[rgb]{0.882,0.882,0.882}}56.1 & {\cellcolor[rgb]{0.882,0.882,0.882}}70.4 & {\cellcolor[rgb]{0.882,0.882,0.882}}80.2 & {\cellcolor[rgb]{0.882,0.882,0.882}}72.3 & {\cellcolor[rgb]{0.882,0.882,0.882}}81.8 & {\cellcolor[rgb]{0.882,0.882,0.882}}42.7 & {\cellcolor[rgb]{0.882,0.882,0.882}}71.6 & {\cellcolor[rgb]{0.882,0.882,0.882}}68.1 & {\cellcolor[rgb]{0.882,0.882,0.882}}77.0 & {\cellcolor[rgb]{0.882,0.882,0.882}}67.7 & 64.6\\
   OW-DETR \cite{gupta2022ow} & 61.8 & 69.1 & 67.8 & 45.8 & 47.3 & 78.3 & 78.4 & 78.6 & 36.2 & 71.5 & {\cellcolor[rgb]{0.882,0.882,0.882}}57.5 & {\cellcolor[rgb]{0.882,0.882,0.882}}75.3 & {\cellcolor[rgb]{0.882,0.882,0.882}}76.2 & {\cellcolor[rgb]{0.882,0.882,0.882}}77.4 & {\cellcolor[rgb]{0.882,0.882,0.882}}79.5 & {\cellcolor[rgb]{0.882,0.882,0.882}}40.1 & {\cellcolor[rgb]{0.882,0.882,0.882}}66.8 & {\cellcolor[rgb]{0.882,0.882,0.882}}66.3 & {\cellcolor[rgb]{0.882,0.882,0.882}}75.6 & {\cellcolor[rgb]{0.882,0.882,0.882}}64.1 & 65.7 \\ 
   \hline
   \textbf{IODC (Ours)} & 66.6 & 75.0 & 62.6 & 49.1 & 53.6 & 75.5 & 83.5 & 76.4 & 43.0 & 73.3 & {\cellcolor[rgb]{0.882,0.882,0.882}}56.1 & {\cellcolor[rgb]{0.882,0.882,0.882}}76.1 & {\cellcolor[rgb]{0.882,0.882,0.882}}84.8 & {\cellcolor[rgb]{0.882,0.882,0.882}}77.9 & {\cellcolor[rgb]{0.882,0.882,0.882}}78.4 & {\cellcolor[rgb]{0.882,0.882,0.882}}42.1 & {\cellcolor[rgb]{0.882,0.882,0.882}}71.2 & {\cellcolor[rgb]{0.882,0.882,0.882}}63.2 & {\cellcolor[rgb]{0.882,0.882,0.882}}72.6 & {\cellcolor[rgb]{0.882,0.882,0.882}}69.2 & \textbf{67.6} \\ 
   \hline
% 10+10 ↑
\multicolumn{1}{l}{} & & & & & & & & & & & & & & & & & & & & \multicolumn{1}{c}{} & \\ 
% 15+5 ↓
   \hline
   \textcolor[RGB]{0, 139, 0}{\textbf{15+5 setting}} & aero & cycle & bird & boat & bottle & bus & car & cat & chair & cow & table & dog & horse & bike & person & {\cellcolor[rgb]{0.863,0.863,0.863}}plant & {\cellcolor[rgb]{0.863,0.863,0.863}}sheep & {\cellcolor[rgb]{0.863,0.863,0.863}}sofa & {\cellcolor[rgb]{0.863,0.863,0.863}}train & {\cellcolor[rgb]{0.863,0.863,0.863}}tv & mAP \\ 
   \hline
   Oracle 20 & 79.4 & 83.3 & 73.2 & 59.4 & 62.6 & 81.7 & 86.6 & 83.0 & 56.4 & 81.6 & 71.9 & 83.0 & 85.4 & 81.5 & 82.7 & {\cellcolor[rgb]{0.882,0.882,0.882}}49.4 & {\cellcolor[rgb]{0.882,0.882,0.882}}74.4 & {\cellcolor[rgb]{0.882,0.882,0.882}}75.1 & {\cellcolor[rgb]{0.882,0.882,0.882}}79.6 & {\cellcolor[rgb]{0.882,0.882,0.882}}73.6 & 75.2 \\
   First 15 & 78.1 & 82.6 & 74.2 & 61.8 & 63.9 & 80.4 & 87.0 & 81.5 & 57.7 & 80.4 & 73.1 & 80.8 & 85.8 & 81.6 & 83.9 & {\cellcolor[rgb]{0.863,0.863,0.863}}- & {\cellcolor[rgb]{0.863,0.863,0.863}}- & {\cellcolor[rgb]{0.863,0.863,0.863}}- & {\cellcolor[rgb]{0.863,0.863,0.863}}- & {\cellcolor[rgb]{0.863,0.863,0.863}}- & 53.2 \\ 
   \hline
   ILOD \cite{shmelkov2017incremental} & 70.5 & 79.2 & 68.8 & 59.1 & 53.2 & 75.4 & 79.4 & 78.8 & 46.6 & 59.4 & 59.0 & 75.8 & 71.8 & 78.6 & 69.6 & {\cellcolor[rgb]{0.863,0.863,0.863}}33.7 & {\cellcolor[rgb]{0.863,0.863,0.863}}61.5 & {\cellcolor[rgb]{0.863,0.863,0.863}}63.1 & {\cellcolor[rgb]{0.863,0.863,0.863}}71.7 & {\cellcolor[rgb]{0.863,0.863,0.863}}62.2 & 65.9 \\
   Faster ILOD \cite{peng2020faster} & 66.5 & 78.1 & 71.8 & 54.6 & 61.4 & 68.4 & 82.6 & 82.7 & 52.1 & 74.3 & 63.1 & 78.6 & 80.5 & 78.4 & 80.4 & {\cellcolor[rgb]{0.863,0.863,0.863}}36.7 & {\cellcolor[rgb]{0.863,0.863,0.863}}61.7 & {\cellcolor[rgb]{0.863,0.863,0.863}}59.3 & {\cellcolor[rgb]{0.863,0.863,0.863}}67.9 & {\cellcolor[rgb]{0.863,0.863,0.863}}59.1 & 67.9 \\
   IODML \cite{joseph2021incremental} & 78.4 & 79.7 & 66.9 & 54.8 & 56.2 & 77.7 & 84.6 & 79.1 & 47.7 & 75.0 & 61.8 & 74.7 & 81.6 & 77.5 & 80.2 & {\cellcolor[rgb]{0.863,0.863,0.863}}37.8 & {\cellcolor[rgb]{0.863,0.863,0.863}}58.0 & {\cellcolor[rgb]{0.863,0.863,0.863}}54.6 & {\cellcolor[rgb]{0.863,0.863,0.863}}73.0 & {\cellcolor[rgb]{0.863,0.863,0.863}}56.7 & 67.8 \\
   ORE \cite{joseph2021towards} & 75.4 & 81.0 & 67.1 & 51.9 & 55.7 & 77.2 & 85.6 & 81.7 & 46.1 & 76.2 & 55.4 & 76.7 & 86.2 & 78.5 & 82.1 & {\cellcolor[rgb]{0.863,0.863,0.863}}32.8 & {\cellcolor[rgb]{0.863,0.863,0.863}}63.6 & {\cellcolor[rgb]{0.863,0.863,0.863}}54.7 & {\cellcolor[rgb]{0.863,0.863,0.863}}77.7 & {\cellcolor[rgb]{0.863,0.863,0.863}}64.6 & 68.5 \\
   OW-DETR \cite{gupta2022ow} & 77.1 & 76.5 & 69.2 & 51.3 & 61.3 & 79.8 & 84.2 & 81.0 & 49.7 & 79.6 & 58.1 & 79.0 & 83.1 & 67.8 & 85.4 & {\cellcolor[rgb]{0.863,0.863,0.863}}33.2 & {\cellcolor[rgb]{0.863,0.863,0.863}}65.1 & {\cellcolor[rgb]{0.863,0.863,0.863}}62.0 & {\cellcolor[rgb]{0.863,0.863,0.863}}73.9 & {\cellcolor[rgb]{0.863,0.863,0.863}}65.0 & 69.1 \\ 
   \hline
   \textbf{IODC (Ours)} & 78.0 & 77.2 & 73.2 & 53.5 & 57.8 & 78.3 & 84.1 & 78.4 & 50.2 & 80.1 & 61.5 & 80.3 & 83.3 & 83.5 & 77.9 & {\cellcolor[rgb]{0.863,0.863,0.863}}41.6 & {\cellcolor[rgb]{0.863,0.863,0.863}}66.2 & {\cellcolor[rgb]{0.863,0.863,0.863}}59.2 & {\cellcolor[rgb]{0.863,0.863,0.863}}70.9 & {\cellcolor[rgb]{0.863,0.863,0.863}}68.4 & \textbf{70.2} \\ 
   \hline
% 15+5 ↑
\multicolumn{1}{l}{} & & & & & & & & & & & & & & & & & & & & \multicolumn{1}{c}{} & \\
% 19+1 ↓
   \hline
   \textcolor[RGB]{0, 139, 0}{\textbf{19+1 setting}} & aero & cycle & bird & boat & bottle & bus & car & cat & chair & cow & table & dog & horse & bike & person & plant & sheep & sofa & train & {\cellcolor[rgb]{0.863,0.863,0.863}}tv & mAP \\ 
   \hline
   Oracle 20 & 79.4 & 83.3 & 73.2 & 59.4 & 62.6 & 81.7 & 86.6 & 83.0 & 56.4 & 81.6 & 71.9 & 83.0 & 85.4 & 81.5 & 82.7 & 49.4 & 74.4 & 75.1 & 79.6 & {\cellcolor[rgb]{0.882,0.882,0.882}}73.6 & 75.2 \\
   First 19 & 76.3 & 77.3 & 68.4 & 55.4 & 59.7 & 81.4 & 85.3 & 80.3 & 47.8 & 78.1 & 65.7 & 77.5 & 83.5 & 76.2 & 77.2 & 46.4 & 71.4 & 65.8 & 76.5 & {\cellcolor[rgb]{0.863,0.863,0.863}} - & 67.5 \\ 
   \hline
   ILOD \cite{shmelkov2017incremental} & 69.4 & 79.3 & 69.5 & 57.4 & 45.4 & 78.4 & 79.1 & 80.5 & 45.7 & 76.3 & 64.8 & 77.2 & 80.8 & 77.5 & 70.1 & 42.3 & 67.5 & 64.4 & 76.7 & {\cellcolor[rgb]{0.863,0.863,0.863}}62.7 & 68.3 \\
   Faster ILOD \cite{peng2020faster} & 64.2 & 74.7 & 73.2 & 55.5 & 53.7 & 70.8 & 82.9 & 82.6 & 51.6 & 79.7 & 58.7 & 78.8 & 81.8 & 75.3 & 77.4 & 43.1 & 73.8 & 61.7 & 69.8 & {\cellcolor[rgb]{0.863,0.863,0.863}}61.1 & 68.6\\
   IODML \cite{joseph2021incremental} & 78.8 & 77.5 & 69.4 & 55.0 & 56.0 & 78.4 & 84.2 & 79.2 & 46.6 & 79.0 & 63.2 & 78.5 & 82.7 & 79.1 & 79.9 & 44.1 & 73.2 & 66.3 & 76.4 & {\cellcolor[rgb]{0.863,0.863,0.863}}57.6 & 70.2 \\
   ORE \cite{joseph2021towards} & 67.3 & 76.8 & 60.0 & 48.4 & 58.8 & 81.1 & 86.5 & 75.8 & 41.5 & 79.6 & 54.6 & 72.8 & 85.9 & 81.7 & 82.4 & 44.8 & 75.8 & 68.2 & 75.7 & {\cellcolor[rgb]{0.863,0.863,0.863}}60.1 & 68.9 \\
   OW-DETR \cite{gupta2022ow} & 70.5 & 77.2 & 73.8 & 54.0 & 55.6 & 79.0 & 80.8 & 80.6 & 43.2 & 80.4 & 53.5 & 77.5 & 89.5 & 82.0 & 74.7 & 43.3 & 71.9 & 66.6 & 79.4 & {\cellcolor[rgb]{0.863,0.863,0.863}}62.0 & 69.8 \\ 
   \hline
   \textbf{IODC (Ours)} & 78.8 & 78.9 & 73.4 & 59.2 & 58.9 & 75.9 & 85.0 & 83.9 & 51.6 & 83.6 & 65.8 & 82.0 & 84.6 & 78.4 & 78.2 & 48.9 & 75.5 & 67.5 & 73.4 & {\cellcolor[rgb]{0.863,0.863,0.863}}62.5 & \textbf{72.3} \\ 
   \hline
   \end{tabular}
\medskip % 迷你字体大小结束符
%\scriptsize
\caption{Comparison of pre-class AP and mAP for incremental object detection on PASCAL VOC 2007. We consider 10+10, 15+5 and 19+1 settings. There are two learning stages in total, and the classes introduced in the second stage are marked with gray. IODML \cite{joseph2021incremental} is the work of incremental detection and the baseline model for our comparison. ORE and OW-DETR are the work of open world object detection. Our method achieves the highest detection performance under three settings.}
\label{tab:1}
\end{table*}
% T1，实验大表格
%-------------------------------------------------------------------------

%-------------------------------------------------------------------------
% 3.2 Text Feature Alignment
%-------------------------------------------------------------------------
\subsection{Text Feature Alignment}
\label{sec:3.2}

The problem with the traditional framework based on Faster-RCNN \cite{ren2015faster} or other detection methods is that its classification head $G$ can only predict the category within closed set. 
As a result, it is not possible to predict the category of an unknown object that has never been seen or labeled. 
Even if we annotate unknown objects with some information, we are still unable to distinguish between sub-classes within an unknown category, which makes it difficult to refine the new classes further.

To overcome the limitation of traditional frameworks, we propose a modification based on CLIP \cite{radford2021learning}. 
As shown in Fig. \ref{fig:1}, we modify the classification output layer in $G$ to a linear mapping layer to obtain the visual embeddings. 
Then, we create a prompt template \cite{gu2021open}, such as `there is a \{classname\} in the scene', where the classname is obtained from $C_{base}$ and $C_{novel}$. 
We then send these prompts containing the class names into CLIP text encoder to obtain the text embeddings, and calculate the cosine similarity between text embeddings and visual embeddings. 
The class name with highest similarity is the predicted category of $G$.

By using the proposed approach, we can overcome the problem of the model not responding to categories other than $C_{base}$ in $T_1$. 
Since CLIP is leverages a corpus of 400 million image-text pairs for training, the correlation between its text embeddings reflects the correlation between visual embeddings. 
Therefore, text embeddings for certain unknown categories can be associated with known category text embeddings through correlations with visual information. 
Thus, aligning with the text embedding space of CLIP helps associate unknown categories with base categories, enhancing the responsiveness to unknown categories. 
Our experiments also reveal that even when the model is trained only using $C_{base}$, it still exhibits small responsiveness to objects $C_{novel}$ category. 
By aligning the incremental model predicted visual embeddings with the CLIP text embeddings, we can enable the model to sense the incoming new classes and reserve their embeddings space to make itself growable for new classes \cite{zhou2022forward}. 

%-------------------------------------------------------------------------
% 3.3 Broad Classes and Category Mapping
%-------------------------------------------------------------------------
\subsection{Broad Classes and Category Mapping}
\label{sec:3.3}

In the incremental learning setting, it is impossible to obtain $I_{novel}$ or even to obtain the exact names of novel classes $C_{novel}$. As a result, using the CLIP text encoder to construct text feature embeddings of $C_{novel}$ is not feasible.

Through our experiments, we have observed that when a CLIP model with zero-shot capability correctly predicts the exact sub-category name, its broad category or parent-category usually ranks second in terms of classification prediction scores. This phenomenon suggests that CLIP can inadvertently establish connections between sub-classes and their parent-classes, such as `cat - animal', `horse - animal', `sofa - furniture', etc. In this context, the term `broad category' refers to manually added classes that typically cover many existing categories. Conversely, the term `unknown category' refers to classes outside the sets $C_i$ in task $T_i$. We refer to these parent classes as broad classes, and we propose introducing them in the $T_{base}$ to address potential unknown classes in the image.

To summarize, the proposed approach involves three main steps. 
In the first step, we use the CLIP text encoder to generate the text embeddings of the base and broad classes and train the model normally in the $T_{base}$. 
In the second step, at the beginning of the $T_{novel}$, we still use the name of the broad classes and observe which broad class is most similar to the newly introduced novel class in the average similarity score through several iterations. 
We then record the one-to-one corresponding relationship between them. However, we cannot convert the learned broad feature to the novel class right away because CLIP's text embeddings are not equivalent for $T_{base}$ and $T_{novel}$. 
Therefore, we continue to use the annotation boxes of novel classes and the text features of the broad classes to learn. Finally, after a period of iterations, we replace the corresponding text features of the broad with correct novel class names to generate new text features and complete the knowledge transfer through this one-to-one category mapping.

%-------------------------------------------------------------------------
% 3.4 Udentify Unknown Classes
%-------------------------------------------------------------------------
\subsection{Identify Unknown Classes}
\label{sec:3.4}

After applying the previous methods, we found that the performance of our model without any reliable supervised signal was still insufficient. Although the model could detect the novel categories, its accuracy was typically lower than 5.0\% of AP50. To address this issue, we further utilized the CLIP image encoder to identify potential unknown objects and improve the model's effectiveness.

To be more precise, in the $T_i$ stage, unknown objects may appear in the proposals if $G$ assigns them to a category other than $C_i$. To address this, we utilize the CLIP image encoder to identify potential objects. If CLIP recognizes the category as belonging to $C_{broad}$ and its prediction score exceeds a predetermined threshold $tr$, we assign the corresponding pseudo broad label to the proposal and train the model using normal cross-entropy classification loss to draw its attention to the broad classes. By incorporating this strategy, which provides explicit supervised information, the model is able to capture unknown objects more effectively compared to our previous approach.

The aforementioned process is executed in real-time during the training phase, which as noted by ViLD \cite{gu2021open}, significantly slows down the entire training process. Moreover, identifying such a large number of proposals using CLIP in a production environment can be quite expensive. Furthermore, the pseudo-labels provided by the CLIP image encoder are insufficient when compared to normal supervised training. As the number of training iterations increases, the RPN layers pay less and less attention to $C_{unk}$, resulting in fewer high-quality proposals being provided to CLIP.

To address the issues of computational cost and pseudo-label quality, we employ NMS on the CLIP-identified unknown proposals before adding them to a pseudo-annotation set. This set serves as a repository for the available unknown categories that are not included in the $C_i$ annotation information for each image. When encountering the same input image again later, we retrieve all the pseudo bounding box information for the same image from the pseudo-annotation set and randomly sample $Z$ bounding boxes to add to the current image annotations for training, where $Z$ is the number of unknown categories in this image. We do not use all the pseudo-annotations because CLIP lacks robustness to images in the detection scene. For an object instance in one image, CLIP often provides multiple prediction instance boxes with high confidence. We randomly sample only a few pseudo-annotations so that the model can deliberately learn the features of the unknown category.

By utilizing the unknown pseudo-annotation set provided by CLIP, the entire network can be improved, resulting in enhanced positioning ability for unknown instances. This method is not employed during testing. Additionally, we utilize knowledge distillation and sampler rehearsal methods, which are commonly used in incremental learning to improve the model's performance for old classes.

%-------------------------------------------------------------------------
% 4. Experiments
%-------------------------------------------------------------------------
\section{Experiments}
\label{sec:4}

We conducted a series of incremental detection experiments on a typical detection dataset to evaluate the effectiveness of our proposed method.

%-------------------------------------------------------------------------
% 4.1 Datasets and Evaluation
%-------------------------------------------------------------------------
\subsection{Datasets and Evaluation}
\label{sec:4.1}

According to the ILOD \cite{shmelkov2017incremental}, we evaluated our proposed method through various incremental detection experiments on the PASCAL VOC 2007 dataset \cite{everingham2009pascal}. The dataset contains 9963 images of 20 categories, with nearly 24k instances. 50\% of the data is used for training and verification, while the rest is used for precision testing.

We use the mean average precision at 0.5 IoU threshold (mAP@50) to evaluate the detection performance of the model. We also pay extra attention to the independent mAP of the model on $C_{base}$ and $C_{novel}$ in different stages for the approach suggested in this research, especially for the accuracy of novel classes, that is, $C_{novel}$.

%-------------------------------------------------------------------------
% T2，消融实验表
%-------------------------------------------------------------------------
\begin{table}
   \centering
   \renewcommand\arraystretch{1.3} % 调整行间距为X倍
   \renewcommand\tabcolsep{2.8pt} % 调整表格列间的宽度
   %\captionsetup{labelformat=empty}
\scriptsize
   \begin{tabular}{l|ccccccccc} 
   \hline
   \multicolumn{1}{l|}{\textcolor[RGB]{0, 139, 0}{\textbf{10+10 setting}}} & \multicolumn{3}{c|}{Ablation Studies} & \multicolumn{3}{c|}{$T_1$} & \multicolumn{3}{c}{$T_2$} \\ 
   \hline
   \multicolumn{1}{l|}{Methods} & text & broad & \multicolumn{1}{c|}{image} & base & novel & \multicolumn{1}{c|}{all} & base & novel & all \\ 
   \hline
   \multicolumn{1}{l|}{Oracle 20} & & & \multicolumn{1}{c|}{} & - & - & \multicolumn{1}{c|}{-} & 74.72 & 75.66 & 75.19 \\
   \multicolumn{1}{l|}{First 10} & & & \multicolumn{1}{c|}{} & 73.40 & 0 & \multicolumn{1}{c|}{36.70} & - & - & - \\ 
   \hline
   \multicolumn{1}{l|}{IODML \cite{joseph2021incremental}} & & & \multicolumn{1}{c|}{} & 75.31 & 0 & \multicolumn{1}{c|}{37.65} & \textbf{68.36} & 64.26 & 66.31 \\
   \multicolumn{1}{l|}{ORE \cite{joseph2021towards}} & & & \multicolumn{1}{c|}{} & - & - & \multicolumn{1}{c|}{-} & 60.37 & 68.79 & 64.58 \\
   \multicolumn{1}{l|}{OW-DETR \cite{gupta2022ow}} & & & \multicolumn{1}{c|}{} & - & - & \multicolumn{1}{c|}{-} & 63.48 & 67.88 & 65.68 \\ 
   \hline%something
   \multirow{4}{*}{\textbf{Ours}} & \checkmark & & \multicolumn{1}{c|}{} & 74.91 & 2.83 & \multicolumn{1}{c|}{38.87} & 66.49 & 65.37 & 65.93 \\
    & \checkmark & \checkmark & \multicolumn{1}{c|}{} & 75.26 & 0.36 & \multicolumn{1}{c|}{37.81} & 66.81 & 65.60 & 66.21 \\
    & & & \multicolumn{1}{c|}{\checkmark} & 75.59 & \textbf{6.88} & \multicolumn{1}{c|}{\textbf{41.23}} & 66.05 & 65.61 & 65.83 \\
    & \checkmark & \checkmark & \multicolumn{1}{c|}{\checkmark} & \textbf{76.95} & 5.36 & \multicolumn{1}{c|}{41.16} & 66.06 & \textbf{69.15} & \textbf{67.61} \\ 
   \hline
% 10+10 ↑
\multicolumn{1}{l}{} & & & & & & & & & \\
% 15+5 ↓
   \hline
   \multicolumn{1}{l|}{\textcolor[RGB]{0, 139, 0}{\textbf{15+5 setting}}} & \multicolumn{3}{c|}{Ablation Studies} & \multicolumn{3}{c|}{$T_1$} & \multicolumn{3}{c}{$T_2$} \\ 
   \hline
   \multicolumn{1}{l|}{Methods} & text & broad & \multicolumn{1}{c|}{image} & base & novel & \multicolumn{1}{c|}{all} & base & novel & all \\ 
   \hline
   \multicolumn{1}{l|}{Oracle 20} & & & \multicolumn{1}{c|}{} & - & - & \multicolumn{1}{c|}{-} & 76.78 & 70.42 & 75.19 \\
   \multicolumn{1}{l|}{First 15} & & & \multicolumn{1}{c|}{} & 76.85 & 0 & \multicolumn{1}{c|}{57.64} & - & - & - \\ 
   \hline
   \multicolumn{1}{l|}{IODML \cite{joseph2021incremental}} & & & \multicolumn{1}{c|}{} & 69.78 & 0 & \multicolumn{1}{c|}{56.26} & 71.73 & 55.90 & 67.77 \\
   \multicolumn{1}{l|}{ORE \cite{joseph2021towards}} & & & \multicolumn{1}{c|}{} & - & - & \multicolumn{1}{c|}{-} & \textbf{73.21} & 58.68 & 68.51 \\
   \multicolumn{1}{l|}{OW-DETR \cite{gupta2022ow}} & & & \multicolumn{1}{c|}{} & - & - & \multicolumn{1}{c|}{-} & 72.21 & 59.84 & 69.12 \\ 
   \hline
   \multirow{4}{*}{\textbf{Ours}} & \checkmark & & \multicolumn{1}{c|}{} & 76.30 & 3.56 & \multicolumn{1}{c|}{58.12} & 73.11 & 59.20 & 69.63 \\
    & \checkmark & \checkmark & \multicolumn{1}{c|}{} & 75.61 & 2.73 & \multicolumn{1}{c|}{57.39} & 72.21 & 57.99 & 68.66 \\
    & & & \multicolumn{1}{c|}{\checkmark} & 76.20 & 6.56 & \multicolumn{1}{c|}{58.79} & 72.81 & 60.93 & 69.84 \\
    & \checkmark & \checkmark & \multicolumn{1}{c|}{\checkmark} & \textbf{76.73} & \textbf{9.21} & \multicolumn{1}{c|}{\textbf{59.85}} & 73.14 & \textbf{61.23} & \textbf{70.16} \\ 
   \hline
% 15+5 ↑
\multicolumn{1}{l}{} & & & & & & & & & \\
% 19+1 ↓
   \hline
   \multicolumn{1}{l|}{\textcolor[RGB]{0, 139, 0}{\textbf{19+1 setting}}} & \multicolumn{3}{c|}{Ablation Studies} & \multicolumn{3}{c|}{$T_1$} & \multicolumn{3}{c}{$T_2$} \\ 
   \hline
   \multicolumn{1}{l|}{Methods} & text & broad & \multicolumn{1}{c|}{image} & base & novel & \multicolumn{1}{c|}{all} & base & novel & all \\ 
   \hline
   \multicolumn{1}{l|}{Oracle 20} & & & \multicolumn{1}{c|}{} & - & - & \multicolumn{1}{c|}{-} & 76.78 & 70.42 & 75.19 \\
   \multicolumn{1}{l|}{First 19} & & & \multicolumn{1}{c|}{} & 71.07 & 0 & \multicolumn{1}{c|}{67.52} & - & - & - \\ 
   \hline
   \multicolumn{1}{l|}{IODML \cite{joseph2021incremental}} & & & \multicolumn{1}{c|}{} & 73.39 & 0 & \multicolumn{1}{c|}{69.72} & 70.89 & 57.60 & 70.23 \\
   \multicolumn{1}{l|}{ORE \cite{joseph2021towards}} & & & \multicolumn{1}{c|}{} & - & - & \multicolumn{1}{c|}{-} & 69.35 & 60.10 & 68.89 \\
   \multicolumn{1}{l|}{OW-DETR \cite{gupta2022ow}} & & & \multicolumn{1}{c|}{} & - & - & \multicolumn{1}{c|}{-} & 70.18 & 62.00 & 69.78 \\ 
   \hline
   \multirow{4}{*}{\textbf{Ours}} & \checkmark & & \multicolumn{1}{c|}{} & \textbf{74.42} & 2.18 & \multicolumn{1}{c|}{\textbf{70.81}} & 72.27 & 59.92 & 71.65 \\
    & \checkmark & \checkmark & \multicolumn{1}{c|}{} & 74.42 & 0.54 & \multicolumn{1}{c|}{70.73} & 72.12 & 58.21 & 71.43 \\
    & & & \multicolumn{1}{c|}{\checkmark} & 73.87 & \textbf{5.26} & \multicolumn{1}{c|}{70.44} & 72.20 & 62.40 & 71.71 \\
    & \checkmark & \checkmark & \multicolumn{1}{c|}{\checkmark} & 74.27 & 3.64 & \multicolumn{1}{c|}{70.74} & \textbf{72.82} & \textbf{62.45} & \textbf{72.30} \\ 
   \hline
   \end{tabular}
\medskip
\caption{Ablation experiment on PASCAL VOC 2007. The comparison is shown in terms of base classes mAP, novel classes mAP and overall mAP. The `text', `broad' and `image' refer to the three methods introduced in Sec. \ref{sec:3}. For the 15+5 setting, base classes contain the first 15 categories, novel classes contain the last 5, and so on.}
\label{tab:2}
\end{table}
% T2，消融实验表
%-------------------------------------------------------------------------

%-------------------------------------------------------------------------
% 4.2 Experimental Settings
%-------------------------------------------------------------------------
\subsection{Experimental Settings}
\label{sec:4.2}

We developed our code using IODML \cite{joseph2021incremental} and followed their experimental settings. As described in Sec. \ref{sec:3.1}, we can only access a subset of the training images $I_i$ for each task $T_i$. If an image contains an instance annotation of a class in $C_i$, it is included in the current training set. If an image does not contain any available annotations, it is removed. 

We arrange 20 category names in PASCAL VOC 2007 in alphabetical order, and divide them into three typical incremental experimental settings following IODML: 10+10, 15+5 and 19+1. The task stages are divided into $T_1$ and $T_2$. 15+5 means that the data of the first 15 base classes can be accessed in $T_1$ and the data of the remaining 5 novels classes can be accessed in $T_2$.

%-------------------------------------------------------------------------
% 4.3 Implementation Details
%-------------------------------------------------------------------------
\subsection{Implementation Details}
\label{sec:4.3}

We built our detector based on Faster-RCNN\cite{ren2015faster}, since the region proposals can be flexibly combined with the CLIP image encoder. We use ResNet-50 \cite{he2016deep} with frozen batch normalization layers as the backbone. 

We have retained the strategies of knowledge distillation and sampler rehearsal in IODML. In the $T_2$ stage, we copied a copy of the frozen parameter model $M_1$ to complete backbone distillation and roi distillation with $M_2$. However, according to \cite{masana2022class}, the choice of sampler rehearsal methods on incremental tasks is not particularly critical. As a result, we disabled the content of the feature store and image store in IODML and used a random sampler instead. Other settings are identical to those in IODML.

We conducted our experiments using four RTX3090s, with a total batch size of eight. We used the SGD optimizer with an initial learning rate of 0.02, reduced to 0.0002 over time, a momentum of 0.9, and a warm-up cycle of 100 iterations.

We leverage CLIP ViT-B/32 \cite{radford2021learning} to generate text embeddings and detect unknown objects. However, the default prompt provided by CLIP, `a photo of a \{classname\}', is not optimized for object detection tasks. Recent open-vocabulary research \cite{gu2021open} has suggested that prompt engineering can improve CLIP's detection performance by better capturing contextual information. Nevertheless, these methods often rely on additional text datasets, such as Flickr30K \cite{plummer2015flickr30k}. In our experiments, we use the prompt `there is a \{classname\} in the scene' without additional datasets. This prompt allows the CLIP text encoder to focus more on environmental information, resulting in better object detection performance.

For the content in Sec. \ref{sec:3.3}, we use several board class names in different incremental tasks, including nouns that have not appeared in the dataset, such as “plant, animal, furniture, vehicle, machine” for 15+5 setting, etc. In the $T_1$ stage, we used $C_{base}$ and $C_{broad}$ to generate text embeddings, while in the $T_2$ stage we obtained the name of the new classes, so we can use $C_{base}$ and $C_{novel}$ to generate new text embeddings. After a period of time in $T_2$, we replace the text embeddings with new ones. Theoretically, adding enough class names would enable CLIP to recognize potential objects belonging to unknown classes and encourage the model to acquire more varied class features. 
However, blindly adding too many broad categories may accidentally cover existing categories. As a result, we employ the same number of board class names as there are novel class names in the experiment. Then, through the category mapping method, we trasfer the knowledge of the broad classes to the new classes.

We discovered that the size of the proposal will vary hugely for the approach proposed in Sec. \ref{sec:3.4}. At times, the size of the proposal is less than $10\times10$ pixels, while the CLIP image encoder predicts a score of more than 0.95. However, manual verification showed that CLIP only matches the most appropriate object in the current class set based on fuzzy texture features, and the predictions are usually incorrect. Therefore, we only identify unknown classes for proposals larger than $100\times100$ pixels and set a prediction threshold of $tr=0.7$ to ensure reliability. It is not critical how many pseudo-annotation boxes are stored. The key is that the CLIP help model points out that there are unknown objects in the image to mitigate the problem of data ambiguity. Even if the position of this box is bad, it is still helpful for learning the following novel classes.

%-------------------------------------------------------------------------
% 4.4 Results
%-------------------------------------------------------------------------
\subsection{Results}
\label{sec:4.4}

Table \ref{tab:1} presents the class-wise average precision (AP) at IoU threshold 0.5 and the corresponding mean average precision (mAP). The novel classes introduced are highlighted in gray. Our method was evaluated in a scenario with two incremental stages and compared with other methods \cite{joseph2021incremental, joseph2021towards, gupta2022ow}.

The second row in each table presents the Oracle model, which represents the upper-bound of accuracy achievable for each class using full supervised learning on all training images. However, different methods have different focuses, and sometimes the average precision (AP) of a class can exceed the Oracle model. In the following lines, we summarize the incremental detection results of three incremental detection works (ILOD \cite{shmelkov2017incremental}, Faster ILOD \cite{peng2020faster}, and IODML \cite{joseph2021incremental}) and two open world detection works (ORE \cite{joseph2021towards} and OW-DETR \cite{gupta2022ow}). It can be observed that our proposed method outperforms the current state-of-the-art in all incremental settings. Our method achieves about a higher mAP of 2\% for all classes in every learning setting compared to other methods.

%-------------------------------------------------------------------------
% T3，K调参表
%-------------------------------------------------------------------------
\begin{table}
   \centering
   \renewcommand\arraystretch{1.3} % 调整行间距为X倍
   \renewcommand\tabcolsep{2.8pt} % 调整表格列间的宽度
\small
   \begin{tabular}{l|ccccccc} 
   \hline
   \multicolumn{1}{l|}{\textcolor[RGB]{0, 139, 0}{\textbf{15+5 setting}}} & \multicolumn{1}{c|}{$K$} & \multicolumn{3}{c|}{$T_1$} & \multicolumn{3}{c}{$T_2$} \\ 
   \cline{3-8}
   \multicolumn{1}{l|}{} & \multicolumn{1}{l|}{values} & base & novel & \multicolumn{1}{c|}{all} & base & novel & all \\ 
   \hline
   \multicolumn{1}{l|}{Ours-text} & \multicolumn{1}{c|}{-} & 76.30 & 3.56 & \multicolumn{1}{c|}{58.12} & 73.11 & 59.20 & 69.63 \\
   \hline
   \multirow{5}{*}{Ours-text} & \multicolumn{1}{c|}{5} & 76.54 & 1.47 & \multicolumn{1}{c|}{57.77} & 72.81 & 60.93 & 69.84 \\
    & \multicolumn{1}{c|}{10} & 76.50 & 2.31 & \multicolumn{1}{c|}{57.95} & 73.33 & 60.39 & 70.10 \\
    & \multicolumn{1}{c|}{15} & 76.28 & 0.84 & \multicolumn{1}{c|}{57.42} & 73.38 & 59.78 & 69.98 \\
    & \multicolumn{1}{c|}{30} & 76.66 & 1.37 & \multicolumn{1}{c|}{57.84} & 73.82 & 58.74 & 70.75 \\ 
    & \multicolumn{1}{c|}{60} & 76.40 & 0.42 & \multicolumn{1}{c|}{57.41} & 73.56 & 57.91 & 69.65 \\ 
   \hline
   \end{tabular}
\medskip
\caption{Effect of adding more redundant category names on results in language space. $K$ is the number of redundant categories added.}
\label{tab:3}
\end{table}
% T3，K调参表
%-------------------------------------------------------------------------

In Table \ref{tab:2}, we present the mAP results of our proposed method for base classes and novel classes at different stages of each experimant. As ORE and OW-DETR are open world works, they cannot provide mAP results in the $T_1$ stage. Therefore, we mainly compared our results with IODML under the same three learning settings. Our method achieves the detection ability of novel classes at an early stage of learning by using text features and broad class names. At the $T_1$ stage, there is a smaller mAP for novel classes, while it is 0 for other methods. ORE and OW-DETR can only detect one unknown class and cannot subdivide other classes from unknown to test. In contrast, our method makes the model more suitable for learning new classes while maintaining backward compatibility. At the $T_2$ stage, our method generally has a higher mAP for novel classes.

We did not include ELI \cite{joseph2022energy} in our comparison as it achieved higher mAP by using a high-level task information during inference to determine whether a latent belongs to the current task or not. This additional signal can indicate the current set of classes, making ELI more like a task-incremental learning approach rather than a class-incremental learning approach. Therefore, it is not directly comparable to the other methods in this paper and our comparison.

%-------------------------------------------------------------------------
% 5 Discussions and Analysis
%-------------------------------------------------------------------------
\section{Discussions and Analysis}
\label{sec:5}

In this section, we analyze the results of our ablation experiments and discuss the motivations behind our design choices. Specifically, we conducted experiments under three different incremental learning settings: 10+10, 15+5, and 19+1.

%-------------------------------------------------------------------------
% 5.1 Ablation Studies
%-------------------------------------------------------------------------
\subsection{Ablation Studies}
\label{sec:5.1}

Table \ref{tab:2} presents the impact of the three proposed methods in Sec. \ref{sec:3} on the mAP in each of these settings. We also report the performance upper-bound of an Oracle model, which is a detector trained on ground-truth annotations for all classes. As expected, the Oracle model achieves the highest performance on all class sets ($C_{base}$, $C_{novel}$, and $C_{all}$).

After implementing the text feature alignment method proposed in Sec. \ref{sec:3.2}, we observed an overall improvement in mAP for both base and novel classes. This suggests that incorporating text features into the visual model can be beneficial, and the language-visual model can parse semantic information from class names. However, after adding the method proposed in Sec. \ref{sec:3.3}, we noticed a slight drop in mAP due to the use of broad class names instead of novel class names. This is because broad classes cover more categories than novel classes. On the other hand, the method proposed in Sec. \ref{sec:3.4} significantly enhances the detection ability of the model for novel classes at $T_1$ by introducing pseudo bounding boxes from the CLIP image encoder. By combining all three methods, we achieved the best results.

These results demonstrate the effectiveness of our proposed methods for incremental detection. By improving the model's detection ability for novel classes in the $T_1$ stage and transferring this knowledge to the $T_2$ stage, we have made the model more forward compatible and enabled it to achieve a higher mAP of novel classes.

Finally, we conducted experiments to verify the importance of text feature embeddings . We added more categories from COCO \cite{lin2014microsoft} under the 15+5 learning setting without using the CLIP image encoder and evaluated the performance in Table \ref{tab:3}. Despite the addition of invalid category names, the results indicate that the language space plays a crucial role in incremental detection. As the number of redundant categories $K$ increases, the overall effect of the experiment remains stable, highlighting the significant impact of language features on the model's detection ability for novel classes. However, when $K \ge 30$, the model's learning ability for novel classes seems to fluctuate.

%-------------------------------------------------------------------------
% F2：蒸馏分析图
%-------------------------------------------------------------------------
\begin{figure}
   \centering %表示居中
   \includegraphics[width=8cm]{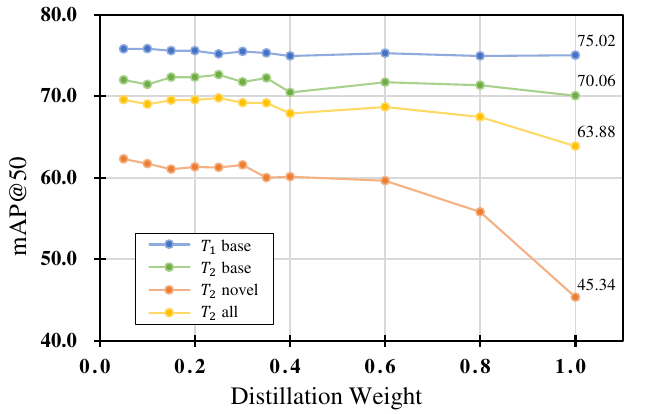}
   % [height=4.5cm]表示高度
   %[width=9.5cm]表示宽度
   %{111.eps}表示eps格式的图片，名为111
   \caption{Experimental results of IODML on PASCAL VOC 2007 under 15+5 setting. With the increase of distillation weight, only the mAP of novel classes in $T_2$ stage has a significant decline.}
   %图片的名称
   \label{fig:2}
   %图片的标签，用于文章中的引用，注意到标签的数字与实际文章显示的数字可能不同
\end{figure}
%-------------------------------------------------------------------------
% F2：蒸馏分析图
%-------------------------------------------------------------------------

%-------------------------------------------------------------------------
% 5.2 Motivation of CLIP
%-------------------------------------------------------------------------
\subsection{Motivation of CLIP}

We have observed that the current incremental detection models are still not able to match the performance of the Oracle model, especially in terms of mAP of new classes. As shown in Table \ref{tab:2}, previous work has focused on maintaining the accuracy of base classes through techniques such as knowledge distillation, resulting in a small gap of within 6\% mAP between the incremental model and the Oracle for of base classes. However, for new classes, the gap is usually greater than 10\% mAP. This indicates a need for further improvement in the accuracy of new classes.

To address this issue, we attempted to modify the weight value of knowledge distillation in IODML to reduce the extent to which the model aligns with the old model of frozen parameters. Fig. \ref{fig:2} shows the results of experiments conducted under the 15+5 setting. The default distillation weight is 0.2. We observed that after the initial learning stage $T_1$, the model was very robust in learning the characteristics of base classes, and even when the distillation weight was reduced, the mAP of base classes did not change much. On the other hand, the mAP of the model for novel classes changed significantly with the distillation weight, since IODML had not deliberately learned the features of novel classes in $T_1$. This finding corroborates the conclusions of other incremental learning work \cite{wu2019large} that the incremental model typically exhibits a strong bias in the last layer of the network and can learn very robust visual features after the first learning stage. In conclusion, modifying the traditional and effective knowledge distillation method did not improve the accuracy of novel classes. Hence, we revisited the data ambiguity problem in incremental detection.

Our motivation is to enhance the model's forward compatibility by enabling it to detect novel classes in the early stages of learning. This approach would enable the model to learn more explicit features and mitigate biases, thereby narrowing the gap with the Oracle. Although open world object detection can achieve this goal, it cannot subdivide multiple unknown classes. To address this, we drew inspiration from open-vocabulary object detection. Typically, a large dataset is required to develop a detection model capable of detecting unknown classes with sufficient object instances that are diverse. Additionally, a massive text dataset is necessary to differentiate between various unknown categories. The potential feature association between text and image would facilitate zero-shot learning for the model.

It is worth noting that the above-mentioned capabilities are highly important for the incremental learning task itself. The current deep learning models often have excess learning capacity on most tasks, which means that introducing additional identified unknown classes for incremental learning tasks would hardly affect the performance of the old classes, and can even often improve the accuracy of all classes. Therefore, having the ability to detect and classify novel classes at an early stage of learning can greatly improve the model's overall performance and make the incremental learning process more efficient and effective. This is why our work focuses on improving the model's ability to detect novel classes from the beginning of the learning process, and our experimental results have shown the effectiveness of our proposed methods for achieving this goal.

Considering the business application background of incremental learning in real life, customers usually only provide us with a limited number of samples at different stages. Obtaining category information beyond customer requirements through expensive manual annotation will certainly help the incremental capability of the model. However, we usually can only get the annotation data of the detection task, and cannot get more text annotation information.

It is almost impossible for the model to obtain an accurate description of the unknown category if it is limited to the current dataset and does not use additional datasets. So we decided to use CLIP \cite{radford2021learning} to help us provide additional information. Although it cannot introduce new datasets, it can offer valuable information to the incremental detection framework in a straightforward and adaptable manner, resulting in a significant improvement in the model's performance. Furthermore, in the future, other more advanced models such as RegionCLIP \cite{zhong2022regionclip} and OpenCLIP \cite{cherti2022reproducible} can be used to further enhance the model's performance.

%-------------------------------------------------------------------------
% 6. Conclusions
%-------------------------------------------------------------------------
\section{Conclusions}

Influenced by the problem of data ambiguity, the current incremental detection method performs well for the base classes and has a weak learning ability for novel classes. We proposed a method to detect the unknown categories to improve the forward compatibility of the model. This method uses CLIP to generate text features for category classification and uses the novel class names to replace the broad class names in the subsequent learning stage, which makes the model obtain better language space globally and can better transfer knowledge from the broad classes to the new classes. Then we use CLIP to detect the potential broad categories in the image and modify these background proposals to target proposals to mitigate the impact of data ambiguity. Our approach effectively improves the model's ability to detect novel categories without relying on additional datasets and outperforms the current state-of-the-art method on benchmark datasets.

%-------------------------------------------------------------------------

{\small
\bibliographystyle{ieee_fullname}
\bibliography{egbib}
}

\end{document}